# EARLY PREDICTION OF ALZHEIMER'S DISEASE DEMENTIA BASED ON BASELINE HIPPOCAMPAL MRI AND 1-YEAR FOLLOW-UP COGNITIVE MEASURES USING DEEP RECURRENT NEURAL NETWORKS

*Hongming Li, Yong Fan*
*for the Alzheimer's Disease Neuroimaging Initiative*

Department of Radiology, Perelman School of Medicine, University of Pennsylvania, Philadelphia, PA, 19104, USA

## ABSTRACT

Multi-modal biological, imaging, and neuropsychological markers have demonstrated promising performance for distinguishing Alzheimer's disease (AD) patients from cognitively normal elders. However, it remains difficult to early predict when and which mild cognitive impairment (MCI) individuals will convert to AD dementia. Informed by pattern classification studies which have demonstrated that pattern classifiers built on longitudinal data could achieve better classification performance than those built on cross-sectional data, we develop a deep learning model based on recurrent neural networks (RNNs) to learn informative representation and temporal dynamics of longitudinal cognitive measures of individual subjects and combine them with baseline hippocampal MRI for building a prognostic model of AD dementia progression. Experimental results on a large cohort of MCI subjects have demonstrated that the deep learning model could learn informative measures from longitudinal data for characterizing the progression of MCI subjects to AD dementia, and the prognostic model could early predict AD progression with high accuracy.

## 1. INTRODUCTION

Alzheimer's disease (AD) is the most prevalent neurodegenerative disorder, and individuals with mild cognitive impairment (MCI) are at a higher risk to develop AD [1]. Although promising performance has been achieved for distinguishing progressive MCI (pMCI) subjects from stable MCI (sMCI) subjects in a pattern classification setting, it remains difficult at baseline to predict which and when MCI individuals will convert to AD dementia.

Neuropsychological tests and a variety of different biological and imaging markers have been explored for diagnosis and following the progression of AD [2-8], such as genetic data, cerebrospinal fluid (CSF) biomarkers, magnetic resonance imaging (MRI), and positron emission tomography (PET). It has been demonstrated that pattern classifiers built upon these measures could distinguish AD patients from subjects with normal cognition (NC) with high accuracy, and recent studies have demonstrated that better performance can be achieved if longitudinal rather than cross-sectional data are used to build the classifiers [9-11].

Longitudinal changes of biomarkers and imaging markers have been investigated for AD diagnosis/prognosis, for instance, changes of neuropsychological measures [12], atrophy rate in cortical thickness and subcortical volume [9, 13-15], and changes in brain tissue intensity/density map [10]. Most longitudinal data based prediction models require different subjects to have data at the same time points. However, missing data is a ubiquitous problem in longitudinal studies. Such a problem is typically circumvented by imputing missing data [11]. Multivariate functional principal component (MFPC) scores [12] have been used to represent longitudinal makers for handling missing or irregular data. However, certain assumption is adopted in MFPC to model the latent longitudinal process, which may not be appropriate for different types of markers.

The early prediction of AD dementia has been typically modeled as a classification problem, for instance, distinguishing sMCI subjects from pMCIs. However, the classification performance is dependent on a cut-off threshold of follow-up duration that is used to define pMCI and sMCI. Moreover, the cohorts of pMCI and sMCI subjects are typically heterogeneous regardless of the threshold used. Furthermore, the classification setting for prediction of AD dementia does not provide timing information about when MCI individuals will cross the threshold to AD dementia. Several studies [11, 12, 16] have focused on the prediction of time of progression to AD under a time-to-event analysis setting, and promising performance have been obtained.

Recently, deep learning techniques built upon recurrent neural networks (RNNs) with a long short term memory (LSTM) [17] structure have achieved remarkable advances in sequence modeling, such as machine translation and functional MRI modeling [18, 19], indicating RNNs might be better tools for characterizing longitudinal data.

In this study, a LSTM autoencoder is adopted to learn compact and informative representation from longitudinal cognitive measures for predicting progression of MCI subjects to AD dementia. These representations could encode temporal dynamics of longitudinal cognitive measures and characterize the progression trajectory of MCI subjects without any explicit assumption regarding the longitudinal process behind the measures. Based on the learned representations and baseline hippocampal MRI data, a prognostic model is built in a time-to-event setting. Particularly, Cox regression model is adopted to estimate the risk of progression of MCI subjects to AD dementia. The proposed model is applied to a large cohort obtained from the Alzheimer's Disease Neuroimaging Initiative (ADNI), and the experimental results have demonstrated that the proposed model could obtain promising prognostic performance, and cognitive measures and imaging based measures could provide complementary information for the prognosis.

## 2. METHODS AND MATERIAL

To build an early prediction model of AD dementia based on longitudinal data, we first train a LSTM autoencoder [20] to learn compact representations and encode temporal dynamics of longitudinal measures for each subject. The learned representations are then combined with baseline imaging data as features to build a prognosis model under a time-to-event analysis setting.

Table1. Demographic information of the dataset used in this study.

| ADNI | | sMCI | pMCI |
|---|---|---|---|
| 1 | Age (years) | 74.92±7.51 | 74.58±7.16 |
| | Gender (M/F) | 114/61 | 132/76 |
| | MMSE | 27.29±1.78 | 26.78±1.74 |
| | Conversion/censor time (months) | 49.2±36.6 | 30.5±25.6 |
| GO&2 | Age | 71.39±7.47 | 72.68±6.97 |
| | Gender (M/F) | 182/150 | 60/47 |
| | MMSE | 28.29±1.62 | 27.35±1.77 |
| | Conversion/censor time (months) | 43.2±16.7 | 24.88±14.5 |
| sMCI: stable MCI who remained as MCI at the last visit; pMCI: progressive MCI who converted to AD before the last visit. | | | |

### 2.1. Data

Cognitive measures of 822 MCI subjects at baseline, 6 months, and 12 months were obtained from ADNI-1, GO & 2, including 13-item version of the Alzheimer's Disease Assessment Scale-Cognitive subscale (ADAS-Cog13), Rey Auditory Verbal Learning Test (RAVLT) immediate, RAVLT learning, Functional Assessment Questionnaire (FAQ), and Mini-Mental State Examination (MMSE). Baseline characteristics of the subjects included are summarized in table 1. Baseline structural MRI scans were also obtained for these subjects to extract hippocampal imaging measures.

### 2.2. LSTM based feature representation

Given the longitudinal cognitive measures at multiple time points for each subject, we learn informative and compact representations to encode the subject's overall longitudinal cognitive performance and its temporal changes/trajectory across multiple time points. The LSTM autoencoder [20] provides an ideal tool for achieving this goal. The architecture of the LSTM autoencoder adopted in this study is illustrated in Fig. 1. It contains two parts, encoder and decoder. The encoder receives the input data of multiple time points and handles the encoding of input measures and their temporal dynamics between consecutive time points. The decoder is utilized to reconstruct the input measures at different time points step by step in a reverse order, based on the learned representations by the encoder. While the network is optimized to minimize the deviation between the reconstructed and input measures, the learned representation by the encoder is expected to characterize the overall cognitive performance and its dynamics of the input longitudinal measures.

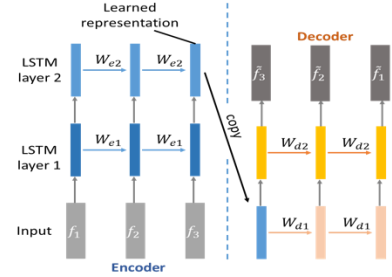

Fig. 1. LSTM autoencoder for longitudinal feature extraction.

As illustrated in Fig. 1, our autoencoder contains 2 LSTM layers for the encoder and decoder respectively. $f_t$ and $\tilde{f}_t$ are the input and reconstructed cognitive measures at time point $t$ ($t$=1,2,3 for illustration), $W_{ei}$ are the trainable parameters of the $i$-th LSTM layer of the encoder, and $W_{di}$ are the trainable parameters of the $i$-th LSTM layer of the decoder. The trainable parameters are those involved in the forget gate, input gate, cell state and hidden state within one LSTM layer. Euclidean distance between the reconstructed and input measures is used as the objective function to optimize the trainable parameters. The number of LSTM layers was chosen to achieve generalizable performance with a small number of trainable parameters.

In this study, the autoencoder of cognitive measures was built on longitudinal cognitive measures of subjects from ADNI-1 cohort. Once the autoencoder was obtained, the encoder was then applied to all the MCI patients from ADNI-1 and GO&2 cohorts to extract their latent features from longitudinal cognitive measures, which were then used for the following prognostic analysis.

### 2.3. Prognostic modeling

Given the latent representations of longitudinal cognitive information, they were combined with baseline hippocampal MRI based measures to build a prognostic model using Cox regression [21].

Particularly, cognitive measures including ADAS-Cog13, RAVLT immediate, RAVLT learning, FAQ, and MMSE were used to learn LSTM encoded cognitive measures. Imaging features from baseline hippocampal MRI data was extracted as an imaging based risk of progression to AD using a deep learning based prognostic framework [22]. Age, gender, education years, and APOEε4 status at baseline were used as covariates in the modeling.

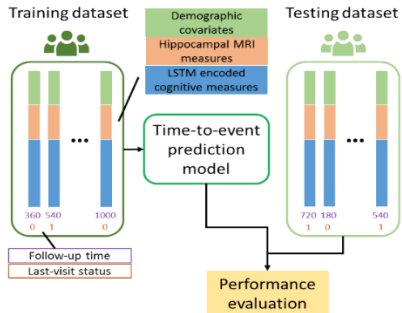

Fig. 2. Schematic diagram of the time-to-event prognosis model.

Cox regression model was built on data of MCI subjects of ADNI-1, and its prognostic performance was evaluated based on the data of MCI subjects of ADNI-GO&2. The schematic illustration of the prognostic modeling is illustrated in Fig. 2. We built prediction models on longitudinal cognitive data of the first 2 and 3 time points separately (within 1-year follow-up) along with the baseline hippocampal imaging based measures.

## 3. RESULTS

### 3.1. Experimental setting

We implemented the LSTM autoencoder using Tensorflow. Two LSTM layers were adopted as shown in Fig. 1, and the number of hidden nodes in each LSTM layer was set to 5 (the dimension of the LSTM encoded cognitive measures, the same as the number of input cognitive measures at each time point). Adam optimization technique was adopted to optimize the autoencoder, with a base learning rate set to 0.01 and updated using a stepwise policy by dropping the learning rate by a factor of 0.1 after every 20000 iterations. The maximum iteration number of the training procedure was set to 100000, and the batch size was set to 64. The autoencoder was trained on a Nvidia Titan Xp graphics processing unit (GPU).

We compared the proposed prediction model with those built on longitudinal cognitive measures only or cognitive measures at single time points. The prognostic performance was evaluated using concordance measure (C-index).

### 3.2. Experimental results

The prognosis performance of the proposed prediction models is demonstrated in Fig. 3. For the LSTM based prediction models built on the longitudinal data, the 6m model was built on data at baseline (bl) and 6-month visits, and the 12m model was built on data at bl, 6-month, and 12-month visits. It is worth noting that if a pMCI subject converted to AD at Visit-ID or data at Visit-ID were not available for a sMCI subject, only data at visits before the Visit-ID were used. When the prediction models built upon ADNI-1 longitudinal cognitive data of 3 and 2 time points applied to ADNI-GO&2 subjects, they achieved C-index values of 0.896 and 0.873 respectively, better than the prediction models built upon single-visit data at 12 and 6 months ($p$=0.019 and 0.552 respectively). All these prediction models had better prediction performance than the model built upon the baseline cognitive measures (a C-index value of 0.848), while the LSTM based prediction models were significantly better ($p<0.05$). When both longitudinal cognitive measures and baseline imaging measures were used, prediction models achieved C-index values of 0.901 and 0.889 at 12 and 6 months respectively, significantly better ($p<0.006$) than a model built on baseline cognitive measures and imaging measures (a C-index value of 0.866).

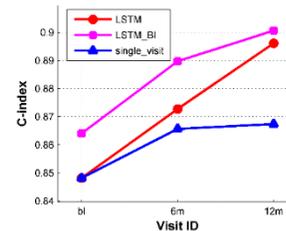

Fig. 3. Prognostic performance of prediction models built on single visit cognitive data (blue), longitudinal cognitive data with 5 LSTM encoded features (red), and longitudinal cognitive data with hippocampal MRI based features (magenta).

As shown in Fig. 3, the prediction models built upon data of any single visit had worse prediction performance than the models built upon longitudinal data, and the models built on data of later time points had better performance than those built on data of earlier time points. Best prognostic performance was obtained when LSTM encoded cognitive representation was combined with imaging based features from baseline hippocampal MRI data [22], indicating that clinical measures and imaging data could provide complementary information for the prognosis. Moreover, we expect that the prognostic performance could be further improved if longitudinal imaging data are incorporated in the prediction model.

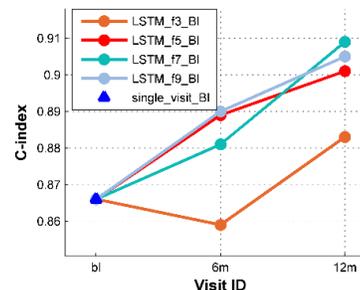

Fig. 4. Prediction performance based on longitudinal data with different numbers of features learned by LSTM autoencoder.

We have also evaluated how the number of hidden nodes in each LSTM layer impacted the prognostic performance. The prognostic performance of prediction models built on longitudinal data with different numbers (3, 5, 7, and 9) of features learned by the LSTM autoencoder and baseline imaging data are shown in Fig. 4, demonstrating that the performance was relatively stable and all the longitudinal data based models outperformed the model built on baseline cognitive and imaging measures when the number of hidden nodes was larger than 3.

## 4. CONCLUSION

In this study, we developed a deep learning based method to characterize longitudinal dynamics of cognitive measures and built prognostic models based on baseline hippocampal MRI measures and the learned longitudinal dynamics to predict individual MCI subjects' progression to AD. Evaluation results have demonstrated that the proposed model achieved promising performance for predicting MCI subjects' progression to AD using data within 1-year follow-up. Future work will be devoted to model optimization and more validation on external data cohorts.

## 5. ACKNOWLEDGEMENTS

This work was supported in part by National Institutes of Health grants [CA223358, EB022573, and AG054409].